%% file: main.tex
\documentclass{article}

\usepackage{iclr2027_preprint,times}

\input{math_commands}

\renewcommand{\eqref}[1]{Eq.~(\ref{#1})}

\usepackage{amsmath,amssymb}

\usepackage{graphicx}
\usepackage{xcolor}
\graphicspath{{figs/}}

\usepackage{booktabs}
\usepackage{threeparttable}
\usepackage{float}
\usepackage{placeins}
\usepackage{multirow}
\usepackage{tabularx}
\usepackage{array}

\usepackage{enumitem}
\usepackage{algorithm}
\usepackage{algpseudocode}

\usepackage[
  colorlinks=true,
  citecolor=blue,
  linkcolor=blue,
  urlcolor=blue
]{hyperref}
\usepackage{url}

\newcommand{\method}{GATS}
\title{Guide, Then Let Go: Gap-Adaptive Teacher Scheduling
for Sparse-Reward Agentic RL}

\author{%
\begin{minipage}[t]{\dimexpr\textwidth-2\tabcolsep\relax}
\centering
\normalfont
\fontsize{10.5}{12.4}\selectfont
\setlength{\parskip}{0pt}
\textbf{Youling Huang}\textsuperscript{1,8,*,$\dagger$}\hspace{0.9em}
\textbf{Tiankuo Xu}\textsuperscript{2,8,*,$\dagger$}\hspace{0.9em}
\textbf{Jiaji Liu}\textsuperscript{3,8,*,$\dagger$}\hspace{0.9em}
\textbf{Tong Zheng}\textsuperscript{4,8,*,$\dagger$}
\\[2pt]
\textbf{Shuo Zhou}\textsuperscript{5,8,$\dagger$}\hspace{1em}
\textbf{Shaotong Qi}\textsuperscript{6,8,$\dagger$}\hspace{1em}
\textbf{Junchi Yao}\textsuperscript{7}\hspace{1em}
\textbf{Shiyang Liu}\textsuperscript{8}
\\[2pt]
\textbf{Hao Xu}\textsuperscript{8}\hspace{1em}
\textbf{Pengcheng Xu}\textsuperscript{8}\hspace{1em}
\textbf{Bo Huang}\textsuperscript{8}\hspace{1em}
\textbf{Hongyi Fu}\textsuperscript{8}\hspace{1em}
\textbf{Lin Lin}\textsuperscript{1,$\ddagger$}
\\[6pt]
\small
\textsuperscript{1}DUT \quad
\textsuperscript{2}XJTU \quad
\textsuperscript{3}THU \quad
\textsuperscript{4}UCAS \quad
\textsuperscript{5}BFSU \quad
\textsuperscript{6}SEU \quad
\textsuperscript{7}MBZUAI \quad
\textsuperscript{8}Kuaishou
\end{minipage}%
}

\hypersetup{
  pdftitle={Guide, Then Let Go: Gap-Adaptive Teacher Scheduling for Sparse-Reward Agentic RL},
  pdfauthor={Youling Huang; Tiankuo Xu; Jiaji Liu; Tong Zheng; Shuo Zhou; Shaotong Qi; Junchi Yao; Shiyang Liu; Hao Xu; Pengcheng Xu; Bo Huang; Hongyi Fu; Lin Lin}
}

\iclrfinalcopy

\IfFileExists{wrapfig2.sty}
  {\usepackage{wrapfig2}}
  {\usepackage{wrapfig}}

\definecolor{placeholdercolor}{RGB}{220,0,0}

\usepackage{listings}
\usepackage[most]{tcolorbox}

\definecolor{promptbackground}{RGB}{248,248,248}
\definecolor{promptborder}{RGB}{180,180,180}
\definecolor{prompttitlebackground}{RGB}{238,238,238}

\newtcblisting{promptbox}[1]{
  enhanced,
  breakable,
  listing only,
  listing engine=listings,
  title={#1},
  colback=promptbackground,
  colframe=promptborder,
  colbacktitle=prompttitlebackground,
  coltitle=black,
  fonttitle=\small\bfseries,
  boxrule=0.45pt,
  arc=1pt,
  outer arc=1pt,
  left=1.8mm,
  right=1.8mm,
  top=1.5mm,
  bottom=1.5mm,
  before skip=5pt,
  after skip=10pt,
  listing options={
    basicstyle=\footnotesize\ttfamily,
    columns=fullflexible,
    breaklines=true,
    breakatwhitespace=true,
    breakindent=1em,
    breakautoindent=false,
    keepspaces=true,
    showstringspaces=false,
    upquote=true
  }
}

\begin{document}

\raggedbottom
\maketitle

\begingroup
\renewcommand{\thefootnote}{\fnsymbol{footnote}}
\footnotetext[1]{Equal contribution.}
\footnotetext[2]{Work done during internships at Kuaishou.}
\footnotetext[3]{Corresponding author.}
\endgroup

\input{sections/00_abstract}

\input{sections/01_introduction}
\input{sections/02_related_work}
\input{sections/03_preliminaries}

\input{sections/04_method}

\input{sections/05_experiments}

\FloatBarrier

\input{sections/06_conclusion}

\bibliography{references}
\bibliographystyle{iclr2027_conference}

\clearpage
\appendix

\input{sections/A_implementation_details}
\input{sections/B_additional_experiments}
\input{sections/C_prompts}

\end{document}

%% file: math_commands.tex
\usepackage{amsmath,amsfonts,bm}

\def\eqref#1{equation~\ref{#1}}

\def\1{\bm{1}}

\DeclareMathAlphabet{\mathsfit}{\encodingdefault}{\sfdefault}{m}{sl}
\SetMathAlphabet{\mathsfit}{bold}{\encodingdefault}{\sfdefault}{bx}{n}



%% file: sections/00_abstract.tex
\begin{abstract}

Reinforcement learning for long-horizon agents typically relies on sparse outcome-based rewards. This leads to a severe cold-start problem, as early-stage policies often fail to solve sampled tasks, leaving little useful reward signal for learning. To mitigate this problem, we use on-policy distillation (OPD) to provide token-level guidance on the student's own rollouts. We find that the benefit of this guidance depends on the performance gap between the teacher and the student. When the teacher substantially outperforms the student, distillation helps guide the student through the early training stage where outcome rewards provide little learning signal. As the gap narrows and eventually reverses, however, continued distillation becomes less beneficial and may hinder further improvement. Motivated by this observation, we propose \textbf{Gap-Adaptive Teacher Scheduling (\method{})}, which augments the student's RL objective with an OPD term whose weight adapts to the teacher--student performance gap. Specifically, \method{} gradually reduces teacher guidance as the student approaches the teacher's reference performance and withdraws it once that reference is reached. This enables \method{} to leverage task-trained teachers smaller than the student, since teacher guidance is primarily needed during early training. Across ALFWorld, WebShop, and ScienceWorld with three Qwen2.5 teacher--student configurations, \method{} achieves the highest average success rate among the compared methods in all three configurations, improving over reward-only GRPO by \textbf{4.37\%--11.87\%} under matched student rollout budgets. Code is available at \url{https://github.com/Ricardo-H/guide-then-let-go}.

\end{abstract}

%% file: sections/01_introduction.tex
\section{Introduction}
\label{sec:introduction}

As model capabilities continue to grow, building agents that can autonomously solve complex long-horizon tasks has become a central question~\citep{zhou2024archer}. Outcome-based reinforcement learning (RL) has become a widely adopted paradigm for training such agents, as it removes the need for a critic network and thereby reduces training complexity~\citep{ji2026tree}. A representative approach is group relative policy optimization (GRPO), which samples a group of trajectories from the current policy and estimates the policy gradient from the relative advantages within the group to maximize the expected return~\citep{shao2024deepseekmath}.

However, in complex long-horizon tasks, GRPO can be limited by the initial policy's inability to discover successful trajectories within a limited rollout budget~\citep{jiang2026mentor,zhang2025bread}. When all trajectories in a sampled group fail and receive the same reward, their group-relative advantages are zero, leaving the group with no reward-driven policy-gradient signal and reducing training efficiency~\citep{zheng2025selective}. This creates a cold-start bottleneck for outcome-based RL, raising the question of how to provide effective guidance when the initial policy cannot discover successful trajectories through exploration alone.

A straightforward way to alleviate this bottleneck is to leverage expert trajectories through supervised fine-tuning before RL~\citep{guo2025deepseekr1} or imitation learning during RL~\citep{zhang2026chord}. In both cases, the expert supervision is off-policy, as the student is trained to increase the token-level likelihood of expert trajectories rather than its own rollouts. The next-token prediction objective enforces rigid, token-level imitation of the expert's trajectory; consequently, the student tends to memorize expert-specific patterns, and the resulting gains transfer poorly beyond the training distribution~\citep{chu2025sftmemorizes}. The problem is further aggravated by the off-policy nature of expert trajectories, as directly fitting them may disrupt the student's established response patterns and induce overfitting to expert data~\citep{zhang2026chord}.

\begin{figure}[t]
    \centering
    \includegraphics[width=\linewidth]{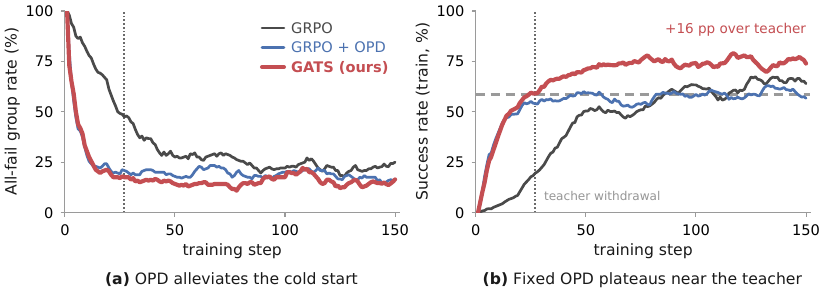}
    \caption{
    \textbf{Guide early, then let go.} Representative training dynamics on WebShop for the
    Qwen2.5-3B$\rightarrow$7B configuration. (a) OPD reduces all-failure rollout groups during the sparse-reward cold start. (b) After teacher withdrawal, \method{} continues to improve beyond the teacher reference, whereas fixed-weight OPD plateaus near it. Dashed lines indicate the teacher reference $M_T$, and dotted lines mark the withdrawal step.
    }
    \label{fig:intro_teaser}
\end{figure}

Another line of work uses expert guidance to steer exploration during RL. The expert may provide a partial prefix for
the model to complete \citep{huang2026prefixrft},
take over generation at designated positions
\citep{jiang2026mentor}, or contribute trajectories to
the model's rollout group \citep{yan2025luffy}. Despite different intervention mechanisms, these approaches inject expert information into the student's exploration process. On-policy distillation (OPD)~\citep{thinkingmachinlab-opd} follows this principle by providing token-level teacher supervision on the student's own rollouts. As illustrated in Figure~\ref{fig:intro_teaser}(a), OPD can substantially reduce the fraction of all-failure groups during early training.

The remaining question is how long such guidance should be maintained. Existing approaches either retain the expert in the training loop or withdraw it according to a manually specified annealing curriculum~\citep{huang2026prefixrft,jiang2026mentor,liu2026uft}.
Such schedules face a trade-off: guidance that is withdrawn too early can leave the model in the sparse-reward regime~\citep{zhang2025bread}, whereas guidance that persists too long, or never fades at all, may restrict the student's later improvement~\citep{li2026sequential}.
This issue is also observed for RL with fixed-weight OPD, which improves rapidly early in training but subsequently plateaus near the teacher reference, as shown in Figure~\ref{fig:intro_teaser}(b).
Thus, expert guidance should adapt to the student's training progress and be withdrawn when it is no longer beneficial.

To address this challenge, we first investigate when OPD is most beneficial and find that its downstream gain correlates strongly and positively with the performance gap between the expert and the student. Motivated by this, we propose \method{}, which augments reinforcement learning with an OPD signal whose weight adapts to this gap. The weight is large early in training, providing stronger guidance when the student struggles to obtain reward-driven learning signals, and decreases as the gap closes. Once the student reaches the teacher reference, the teacher is withdrawn and training continues with GRPO alone. Because the expert is needed only for this early directional guidance, a small expert model suffices, which substantially reduces distillation cost. Experiments on ALFWorld, WebShop, and ScienceWorld with three Qwen2.5 teacher--student configurations show that, at the same training budget, \method{} improves over GRPO by 4.37\%--11.87\%. Gap-adaptive on-policy distillation guides the student through the cold-start stage, where outcome rewards provide little gradient, and its automatic withdrawal leaves room for free exploration, so the student can surpass its teacher instead of merely converging to it. Ablations attribute these gains to the adaptive schedule itself rather than to distillation alone.

Our contribution can be summarized as follows:

\begin{itemize}[leftmargin=*]  \item Through experiments, 
we identify a strong monotone relationship between the gain from on-policy 
distillation and the teacher--student performance gap, including a sign 
reversal at capability crossover: token-level guidance accelerates learning 
while the teacher is ahead, but actively suppresses the student once the 
gap closes.
  \item Building on this observation, we propose \method{}, which augments the RL 
objective with an OPD term whose weight is an adaptive, monotone function 
of the measured performance gap and vanishes at crossover. \method{} 
requires neither imitation of fixed expert trajectories nor a hand-designed 
annealing schedule. Moreover, since the teacher is only needed for early 
directional guidance, it can be smaller than the student, which 
substantially reduces the cost of distillation.
    \item We conduct experiments on three benchmarks (ALFWorld, WebShop, and ScienceWorld) under multiple teacher--student configurations, showing that \method{} consistently outperforms strong baselines under the same student rollout budget. Ablations further verify that the gains come from the gap-adaptive schedule.
\end{itemize}

%% file: sections/02_related_work.tex
\section{Related Work}
\label{sec:related-work}

\paragraph{Reinforcement Learning for Agentic LLMs.} Reinforcement learning has been increasingly adopted to enhance the agentic capabilities of LLMs, encompassing hierarchical planning~\citep{zhou2024archer}, tool invocation~\citep{feng2026retool}, and multi-turn interaction with external environments~\citep{jin2025search}. Many of these methods rely on automatically verifiable feedback from the environment, such as signals of task completion~\citep{wang2025ragen}. However, agentic RL is highly sensitive to the initial competence of the policy: stronger pretrained priors yield higher initial rewards and thereby enable more effective policy improvement, whereas weaker agents often struggle to obtain successful trajectories in long-horizon environments~\citep{bai2024digirl}. In this cold-start regime, sparse outcome rewards cause most sampled trajectories to fail; consequently, the within-group reward variance, and hence the group-relative advantage, can collapse to zero, leaving many costly interactive rollouts without an effective learning signal~\citep{xi2025agentgym,yu2026dapo}.

\paragraph{Combining RLVR with OPD.} Recent work has begun to use teacher signals from on-policy distillation~\citep{agarwal2024policy}  to compensate for the sparse outcome rewards in RLVR~\citep{shao2024deepseekmath,yu2026dapo}. One line of research decides which samples should receive OPD supervision based on external information, for example applying it to incorrect groups~\citep{li2026unifying}, to groups where all rollouts fail~\citep{ding2026hdpo}, or to groups with large teacher--student disagreement~\citep{zhong2026sod}. Another line studies how teacher supervision should evolve over the course of training, for example linearly annealing the teacher signal to gradually reduce its influence~\citep{tan2026atod}. However, these methods either retain the expert in the training loop throughout the entire training process~\citep{zhang2026chord}, which prevents the student from surpassing the teacher ceiling~\citep{li2026sequential}, or rely on manually predefined annealing schedules~\citep{tan2026atod}. How to anneal the teacher signal adaptively according to the actual training dynamics, and to eventually withdraw it so that the student can surpass the teacher, remains an open problem.

%% file: sections/03_preliminaries.tex
\section{Preliminaries}
\label{sec:preliminary}

\subsection{Agentic Setting}

Let $\mathcal{D}$ denote the distribution of training prompts, where each prompt $\mathbf{x}\sim\mathcal{D}$ specifies an agentic task. For each prompt $\mathbf{x}$, the student model samples a group of $G$ trajectories, denoted by $\{\mathbf{y}_i\}_{i=1}^{G}$. Each trajectory is represented as $\mathbf{y}_i=(y_{i,1},\ldots,y_{i,|\mathbf{y}_i|})$, where $y_{i,n}$ denotes the token generated at position $n$ and $|\mathbf{y}_i|$ is the trajectory length. The conditioning context for $y_{i,n}$ is denoted by $\mathbf{h}_{i,n}$, which comprises the prompt, previously generated tokens, intermediate actions, and observed environment feedback available before generating $y_{i,n}$. Each trajectory $\mathbf{y}_i$ receives an outcome reward $r_i$ determined by the task-specific evaluation criterion.

\subsection{Group Relative Policy Optimization}
\label{sec:prelim-grpo}

GRPO optimizes the student policy using
relative outcome feedback within each sampled group \citep{shao2024deepseekmath}.
Specifically, each trajectory's
reward is centered by the group mean and scaled by the group standard deviation, yielding the advantage
\begin{equation}
  \widehat{A}_i
  = \frac{r_i-\overline{r}}{\sigma_r+\epsilon},
  \qquad
  \overline{r}=\frac{1}{G}\sum_{j=1}^{G}r_j,
  \label{eq:grpo-advantage}
\end{equation}
where $\overline{r}$ and $\sigma_r$ are the mean and
standard deviation of the within-group reward, respectively, and $\epsilon>0$ is a small constant.
The importance ratio is
\begin{equation}
  \rho_{i,n}(\theta)
  =
  \frac{\pi_\theta(y_{i,n}\mid\mathbf{h}_{i,n})}
       {\pi_{\theta_{\mathrm{old}}}(y_{i,n}\mid\mathbf{h}_{i,n})},
  \label{eq:grpo-ratio}
\end{equation}
where $\pi_{\theta_{\mathrm{old}}}$ and $\pi_\theta$ denote the old and current
student policies, respectively. With clipping threshold $\varepsilon>0$, the
GRPO objective is
\begin{equation}
\begin{split}
  \mathcal{L}_{\mathrm{GRPO}}
  = -\mathbb{E}\Bigg[
  \frac{1}{G}\sum_{i=1}^{G}\frac{1}{|\mathbf{y}_i|}\sum_{n=1}^{|\mathbf{y}_i|}
  \min\Bigg(
    \rho_{i,n}(\theta)\widehat{A}_i,\,
    \operatorname{clip}\!\left(
      \rho_{i,n}(\theta),1-\varepsilon,1+\varepsilon
    \right)\widehat{A}_i
  \Bigg)
  \Bigg].
  \label{eq:grpo-loss}
\end{split}
\end{equation}

\begin{figure}[t]
    \centering
    \includegraphics[width=0.98\linewidth]{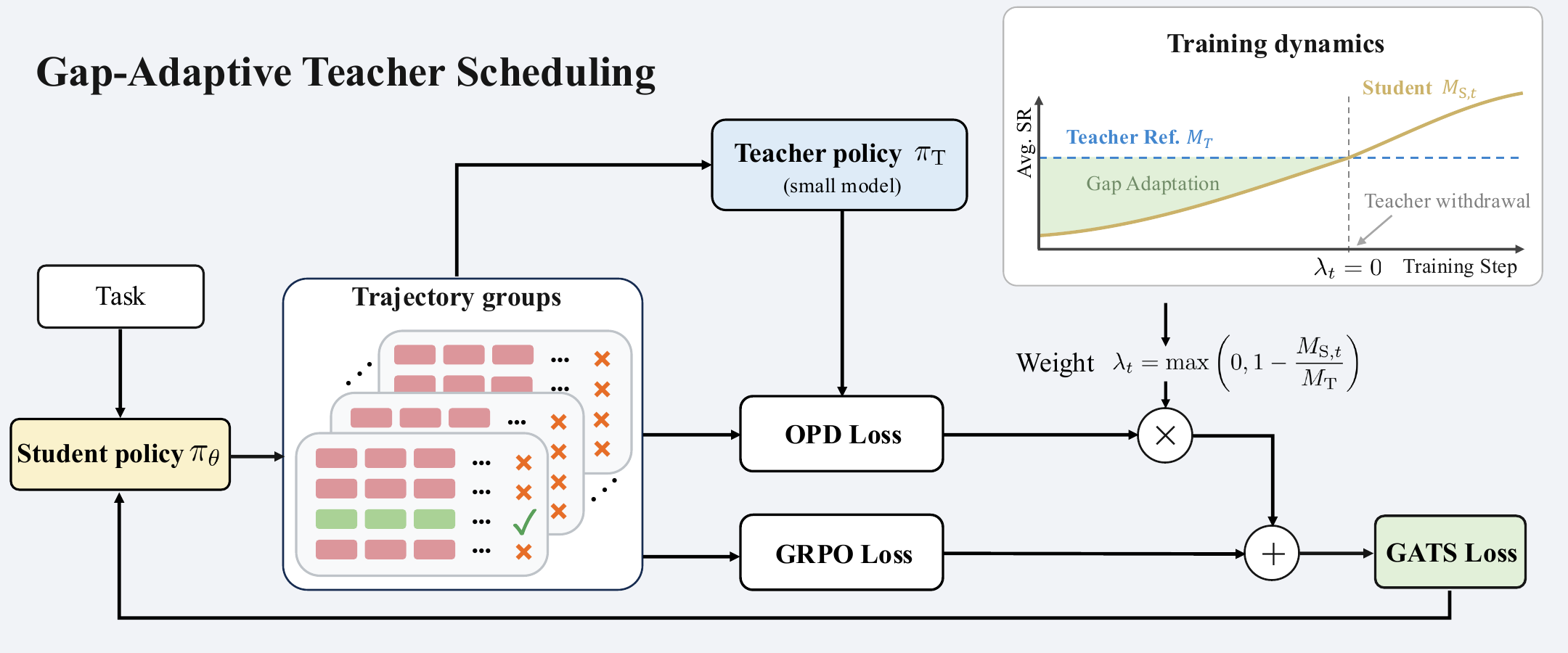}
    \caption{\textbf{Overview of \method{}.}
    The student generates trajectory groups that are used to compute both the GRPO loss and the teacher-guided OPD loss. GATS adaptively weights the OPD loss according to the teacher--student capability gap and combines it with the GRPO loss to update the student. As the gap narrows, the OPD weight decreases to zero, after which the teacher is permanently withdrawn and training continues with GRPO alone.}
    \label{fig:method-overview}
\end{figure}
\subsection{On-Policy Distillation}
\label{sec:prelim-opd}

OPD provides dense token-level supervision on
student-generated trajectories by aligning the student policy with a frozen
teacher policy \citep{agarwal2024policy}. For each sampled token $y_{i,n}$, the OPD advantage
is defined as
\begin{equation}
  A^{\mathrm{OPD}}_{i,n}
  =
  \log\pi_{\mathrm{T}}(y_{i,n}\mid\mathbf{h}_{i,n})
  -
  \log\pi_\theta(y_{i,n}\mid\mathbf{h}_{i,n}),
  \label{eq:opd-advantage}
\end{equation}
where the teacher log-probabilities are detached from the gradient computation.
The corresponding OPD objective is
\begin{equation}
  \mathcal{L}_{\mathrm{OPD}}
  =
  -\mathbb{E}\left[
    \frac{1}{G}\sum_{i=1}^{G}
    \frac{1}{|\mathbf{y}_i|}
    \sum_{n=1}^{|\mathbf{y}_i|}
    \rho_{i,n}(\theta) A^{\mathrm{OPD}}_{i,n}
  \right],
  \label{eq:opd-loss}
\end{equation}
where $\rho_{i,n}(\theta)$ is the importance ratio defined in
Eq.~\ref{eq:grpo-ratio}.

%% file: sections/04_method.tex
\section{Gap-Adaptive Teacher Scheduling}
\label{sec:method}

\begin{wrapfigure}{r}{0.48\linewidth}
    \vspace{-\intextsep}
    \centering
    \includegraphics[width=\linewidth]{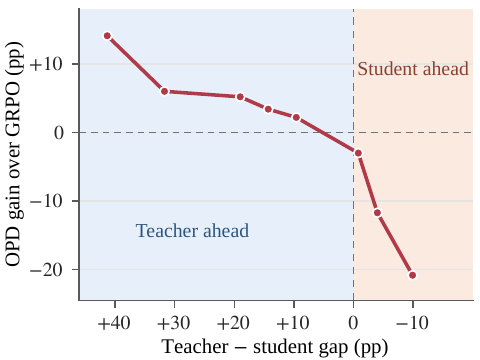}
    \caption{OPD gain over GRPO versus the teacher--student performance
    gap. The observed gain diminishes as the gap narrows and becomes negative near parity, motivating gap-adaptive supervision and eventual
    teacher withdrawal.}
    \label{fig:prelim-gap-opd}
    \vspace{-\intextsep}
\end{wrapfigure}

\paragraph{Overview.}
Figure~\ref{fig:method-overview} illustrates Gap-Adaptive Teacher Scheduling (GATS). A teacher policy $\pi_{\mathrm{T}}$ is first trained on the target tasks, and its late-stage performance is used to define a fixed teacher reference score $M_{\mathrm{T}}$. During student training, each group of student-generated trajectories is used to compute both the outcome-driven GRPO objective in \eqref{eq:grpo-loss} and the teacher-guided OPD objective in \eqref{eq:opd-loss}. Motivated by the gap--utility trend in Figure~\ref{fig:prelim-gap-opd}, GATS scales the OPD objective by an adaptive coefficient $\lambda_t$ determined by the gap between the moving-average student score $M_{\mathrm{S},t}$ and the teacher reference score $M_{\mathrm{T}}$. As this gap narrows, $\lambda_t$ gradually decreases, reducing the contribution of teacher guidance to the GATS objective. Once $M_{\mathrm{S},t}\geq M_{\mathrm{T}}$, the teacher branch is permanently withdrawn, and subsequent training proceeds with GRPO alone. The full procedure is summarized in Algorithm~\ref{alg:phase-out}.

\paragraph{Empirical motivation.}
To characterize the marginal utility of OPD, we fix an RL-enhanced Qwen2.5-1.5B teacher and select several increasingly capable Qwen2.5-3B student checkpoints. Each student checkpoint is trained
on ALFWorld for 15 steps under the same budget, using either GRPO alone or GRPO
with OPD, and we compare the resulting change in success rate.
Figure~\ref{fig:prelim-gap-opd} shows that the benefit of OPD decreases as the
gap narrows and becomes negative once the student exceeds the teacher in
capability. This result motivates adapting the OPD weight to the current
teacher--student capability gap rather than keeping it fixed throughout
training.

\paragraph{Teacher reference score.}
Let $T_{\mathrm{T}}$ be the total number of teacher training steps, and
$m_{\mathrm{T},t}$ its training-set score at step $t$. Given a window size $K$,
we define the teacher reference score as the average over the last $K$
measurements:
\begin{equation}
\label{eq:teacher-reference}
    M_{\mathrm{T}}
    = \frac{1}{K}\sum_{t=T_{\mathrm{T}}-K+1}^{T_{\mathrm{T}}} m_{\mathrm{T},t}.
\end{equation}
This scalar is fixed throughout student training and represents the capability
level at which teacher guidance is no longer needed.

\paragraph{Student capability estimate.}
Similarly, we track the student's training-set score $m_{\mathrm{S},t}$ at each
step $t$. To prevent the current update from affecting its own distillation
weight, we estimate the student's capability using a one-step-lagged moving
average:
\begin{equation}
\label{eq:student-trailing}
    M_{\mathrm{S},t}
    =
    \frac{1}{|\mathcal{H}_t|}
    \sum_{k\in\mathcal{H}_t}m_{\mathrm{S},k},
    \qquad
    \mathcal{H}_t
    =
    \{k:\max(0,t-K)\leq k\leq t-1\}.
\end{equation}
When fewer than $K$ previous measurements are available, the average is computed
over all available historical measurements.

\paragraph{Teacher withdrawal.}
The OPD weight is determined by the normalized gap between the teacher reference
score and the lagged student score:
\begin{equation}
\label{eq:lambda}
    \lambda_t
    = \max\!\left(1 - \frac{M_{\mathrm{S},t}}{M_{\mathrm{T}}},\,0\right).
\end{equation}
Thus, $\lambda_t$ is large when the student is far below the teacher reference
level and decreases as the student approaches that level. Teacher guidance is
withdrawn once
\begin{equation}
\label{eq:deactivation-condition}
    M_{\mathrm{S},t} \ge M_{\mathrm{T}}.
\end{equation}
Let $d_t\in\{0,1\}$ denote the withdrawal flag at step $t$. It is initialized
as $d_0=0$ and is set to one after \eqref{eq:deactivation-condition} is
satisfied. When $d_t=1$, the teacher forward pass is skipped, and the OPD weight
is set to zero.

\paragraph{Adaptive training objective.}
At student training step $t$, \method{} optimizes the following objective:
\begin{equation}
\label{eq:full-loss}
    \mathcal{L}_{\mathrm{GATS},t}
    =
    \mathcal{L}_{\mathrm{GRPO}}
    +
    (1-d_t)\lambda_t
    \mathcal{L}_{\mathrm{OPD}}.
\end{equation}
Before teacher withdrawal, $d_t=0$, and the GATS loss combines GRPO with
gap-adaptive OPD. Once $M_{\mathrm{S},t}\geq M_{\mathrm{T}}$, we set $d_t=1$, withdraw the teacher by setting the OPD
term to zero, and continue training with GRPO alone. The resulting procedure provides
dense teacher guidance while the student remains below the reference capability
and gradually reduces this guidance as the student's capability approaches that
of the teacher.

%% file: sections/05_experiments.tex
section{Experiments}
\subsection{Experimental Setup}

\paragraph{Benchmarks and evaluation.}
We evaluate on ALFWorld~\citep{shridhar2020alfworld} for household
instruction following, WebShop~\citep{yao2022webshop} for online shopping,
and ScienceWorld~\citep{wang2022scienceworld} for scientific experimentation.
We report success rates (SR) on the in-distribution (ID) and
out-of-distribution (OOD) splits of ALFWorld and ScienceWorld, and on
the WebShop evaluation split.
Avg.~SR is the unweighted mean of these five benchmark--split metrics.
For each trained policy, we evaluate the final checkpoint three times,
using 128 episodes per evaluation, and report the mean SR.
Prompt-only models follow the same evaluation protocol.
The three repetitions use a fixed checkpoint.
Benchmark splits, interaction protocols, and evaluation details are
provided in Appendix~\ref{app:env-protocol}.

\paragraph{Models and training.}
All teachers and students are initialized from the Qwen2.5-Instruct
family~\citep{Yang2024Qwen25TR}.
We consider three teacher--student configurations:
$1.5\mathrm{B}\!\rightarrow\!7\mathrm{B}$,
$1.5\mathrm{B}\!\rightarrow\!14\mathrm{B}$, and
$3\mathrm{B}\!\rightarrow\!7\mathrm{B}$.
Each teacher is trained with GRPO on the corresponding environment
and then frozen during student training.
All trained student methods are run for 150 updates.
Within each environment and student size, we hold the training data,
student rollout budget, and common optimization hyperparameters fixed
across methods.
Model and training configurations are detailed in
Appendix~\ref{app:training-config}.

\paragraph{Baselines.}
We compare \method{} with five baselines.
\textbf{Prompt-only} evaluates the instruction-tuned model without
additional training.
\textbf{GRPO} uses outcome-driven RL without teacher
supervision~\citep{shao2024deepseekmath}.
In our implementation, the three distillation-based baselines
combine GRPO with OPD:
\textbf{GRPO $+$ OPD} uses a fixed OPD weight;
\textbf{ATOD} reduces the OPD weight according to a predefined
annealing schedule~\citep{tan2026atod}; and
\textbf{SOD} adjusts teacher supervision according to
teacher--student divergence while retaining it throughout
training~\citep{zhong2026sod}.
Baseline implementations and method-specific hyperparameters are
provided in Appendix~\ref{app:baselines}.

\begin{table}[t]
\centering
\caption{Final success rates (\%) on ALFWorld, WebShop, and ScienceWorld.
Each entry is averaged over three evaluation runs.
Avg.~SR denotes the unweighted mean of the five reported metrics.
Within each student block, the best and second-best results in each column are
shown in bold and underlined, respectively.}
\label{tab:main-results}
\small
\setlength{\tabcolsep}{3pt}
\renewcommand{\arraystretch}{0.95}
\begin{tabular*}{\linewidth}{@{\extracolsep{\fill}}lrrrrrr@{}}
\toprule
& \multicolumn{2}{c}{\textbf{ALFWorld}}
& \multicolumn{1}{c}{\textbf{WebShop}}
& \multicolumn{2}{c}{\textbf{ScienceWorld}}
& \multicolumn{1}{c}{\textbf{Avg.~SR}} \\
\cmidrule(lr){2-3}
\cmidrule(lr){4-4}
\cmidrule(lr){5-6}
\cmidrule(l){7-7}
Method & ID & OOD & Eval & ID & OOD & \\
\midrule

\textbf{Teacher: Qwen2.5-1.5B}
  & 53.65 & 60.42 & 63.80 & 12.76 & 13.80 & 40.89 \\

\midrule
\multicolumn{7}{@{}l@{}}{\textit{Student: Qwen2.5-7B}} \\
\addlinespace[1pt]

Prompt-only
  & 14.84 & 13.02 & 0.26 & 10.94 & 7.03 & 9.22 \\

GRPO
  & 63.02 & \underline{73.70} & 61.20
  & \underline{38.02} & \underline{28.91}
  & \underline{52.97} \\

GRPO $+$ OPD
  & 55.21 & 46.09 & 61.20
  & 18.49 & 14.58 & 39.11 \\

ATOD
  & \underline{75.00} & 56.51 & \underline{68.49}
  & 27.34 & 21.09 & 49.69 \\

SOD
  & 59.38 & 59.38 & 64.32
  & 17.19 & 10.42 & 42.14 \\

\textbf{\method{}}
  & \textbf{84.38} & \textbf{78.12} & \textbf{76.30}
  & \textbf{48.44} & \textbf{36.98}
  & \textbf{64.84} \\

\midrule
\multicolumn{7}{@{}l@{}}{\textit{Student: Qwen2.5-14B}} \\
\addlinespace[1pt]

Prompt-only
  & 44.01 & 52.34 & 1.30
  & 25.26 & 28.65 & 30.31 \\

GRPO
  & \underline{71.09} & \underline{74.22} & \underline{70.05}
  & \underline{51.56} & \textbf{42.71}
  & \underline{61.93} \\

GRPO $+$ OPD
  & 52.60 & 54.69 & 64.58
  & 17.45 & 15.36 & 40.94 \\

ATOD
  & 63.54 & 66.93 & 66.67
  & 19.53 & 21.61 & 47.66 \\

SOD
  & 53.39 & 52.34 & 65.36
  & 13.80 & 12.24 & 39.43 \\

\textbf{\method{}}
  & \textbf{82.81} & \textbf{79.95} & \textbf{74.48}
  & \textbf{52.08} & \underline{42.19}
  & \textbf{66.30} \\

\midrule

\textbf{Teacher: Qwen2.5-3B}
  & 69.01 & 69.53 & 50.00
  & 35.68 & 29.69 & 50.78 \\

\midrule
\multicolumn{7}{@{}l@{}}{\textit{Student: Qwen2.5-7B}} \\
\addlinespace[1pt]

Prompt-only
  & 14.84 & 13.02 & 0.26
  & 10.94 & 7.03 & 9.22 \\

GRPO
  & 63.02 & 73.70 & 61.20
  & 38.02 & 28.91 & 52.97 \\

GRPO $+$ OPD
  & 69.79 & 67.71 & 57.55
  & 32.81 & 28.12 & 51.20 \\

ATOD
  & \underline{75.00} & \textbf{78.65} & \underline{71.09}
  & \underline{44.53} & \underline{36.72}
  & \underline{61.20} \\

SOD
  & 65.89 & 72.66 & 60.42
  & 38.80 & 28.12 & 53.18 \\

\textbf{\method{}}
  & \textbf{76.82} & \underline{76.56} & \textbf{78.39}
  & \textbf{50.52} & \textbf{41.41}
  & \textbf{64.74} \\

\bottomrule
\end{tabular*}
\end{table}

\subsection{Overall Performance}

\paragraph{\method{} achieves consistent performance gains across all teacher-student configurations.} As shown in Table~\ref{tab:main-results}, \method{} attains the highest average success rate under all three teacher-student configurations. With a Qwen2.5-1.5B teacher and a Qwen2.5-7B student, it improves the average success rate of outcome-reward-only GRPO from 52.97\% to 64.84\% (+11.87 points); comparable improvements of +4.37 and +11.77 points are observed in the settings with a Qwen2.5-14B student and a Qwen2.5-3B teacher, respectively. Notably, \method{} also yields improvements on the OOD test sets of ALFWorld and ScienceWorld, indicating that the capabilities acquired under teacher guidance generalize beyond the training distribution. These results demonstrate that \method{} effectively alleviates the cold-start problem in long-horizon agent training.

\paragraph{Fixed or predefined schedules for teacher supervision cannot adapt to the dynamically learning progress.} The results in Table~\ref{tab:main-results} show that teacher supervision can fail in two opposite directions. On the one hand, supervision that is never withdrawn anchors the student to the capability ceiling of the teacher: GRPO + OPD and SOD perform even worse than plain GRPO, since neither of them fully withdraws the influence of the teacher. On the other hand, the annealing process in ATOD follows a predefined schedule that is decoupled from the actual training dynamics: withdrawing supervision too early re-exposes the student to the sparse-reward dilemma, whereas withdrawing it too late pulls the student toward the suboptimal teacher policy. Consequently, the performance of ATOD fluctuates sharply across configurations, with average success rates ranging from 47.66\% to 61.20\%. In contrast, \method{} ties the supervision strength directly to the measured performance gap and withdraws supervision entirely once the student reaches the reference performance of the teacher, thereby achieving the best results across all configurations. These observations support our central claim: teacher intervention must track the actual progress of training and be withdrawn adaptively.

\subsection{Training Dynamics: From Cold Start to Teacher Withdrawal}
\label{sec:mechanism-analysis}

\begin{figure}[t]
    \centering
    \includegraphics[width=\linewidth]{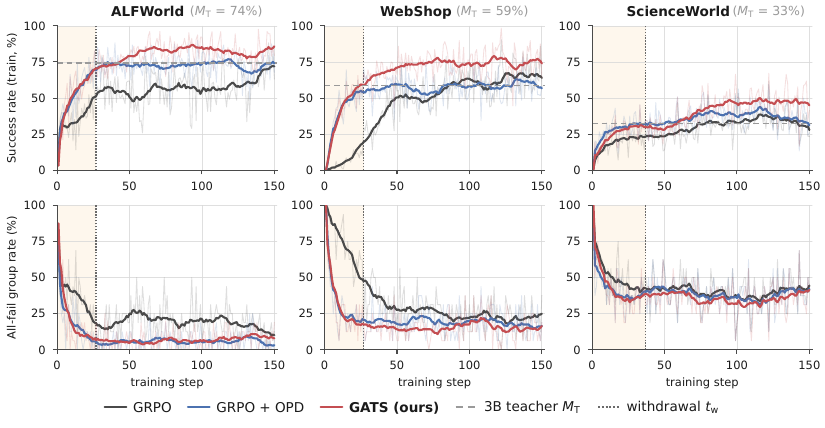}
    \caption{
    Training dynamics for Qwen2.5-3B$\rightarrow$7B on ALFWorld, WebShop, and ScienceWorld.
    Top: training success rate. Bottom: all-failure group rate.
    Horizontal dashed lines denote the teacher reference $M_T$; vertical dotted lines mark GATS teacher withdrawal at $t_w$.
    Teacher guidance accelerates early learning, while \method{} continues to improve after withdrawal.
    }
    \label{fig:training-dynamics}
\end{figure}

\paragraph{Early guidance from the teacher effectively mitigates the cold-start problem.} As shown in Figure~\ref{fig:training-dynamics}, during the early stage of training (shaded region), both GRPO+OPD and \method{} achieve substantially higher rollout success rates than vanilla GRPO, along with a markedly lower proportion of all-failure groups. Unlike vanilla GRPO, the two distillation-based methods receive token-level supervision signals from the very first training step, and consequently their all-failure rates decrease rapidly. These results confirm that OPD yields the largest gains precisely at the stage where successful trajectories are scarcest and the outcome reward is least informative.

\paragraph{Persistent teacher supervision that is never withdrawn anchors the student near the reference level of the teacher.} The early advantage of fixed-weight GRPO+OPD gradually diminishes as training proceeds. Its success rates plateau around the teacher reference values
$M_T=0.74$, $0.59$, and $0.33$ for ALFWorld, WebShop, and ScienceWorld, respectively. Despite its slower start, pure GRPO eventually catches up to GRPO+OPD in all three environments. This observation indicates that supervision without withdrawal confines the student below the capability ceiling of the teacher, such that the acceleration gained in the early phase is entirely offset by the ceiling effect in the later phase.

\paragraph{\method{} achieves both early-stage acceleration and late-stage breakthroughs, as learning continues even after teacher withdrawal.} Specifically, \method{} permanently withdraws the teacher once the estimated student competence reaches the threshold, at the withdrawal step $t_w$ marked by the dashed line in the figure. After teacher withdrawal, the rollout success rate continues to improve and eventually exceeds the corresponding teacher reference in all three environments. These training dynamics directly validate the two-stage design of \method{}: the teacher provides dense guidance when it is most needed and is withdrawn once the student no longer requires it, leaving the subsequent reward-driven learning entirely unconstrained.

\subsection{Annealing Schedule Comparison}
\label{sec:annealing-comparison}

\begin{table}[t]
\centering
\begin{minipage}{0.82\linewidth}
\centering
\caption{Comparison of OPD-weight decay schedules with a Qwen2.5-1.5B
teacher and Qwen2.5-7B student.
ALF and Sci.\ denote the ID splits of ALFWorld and ScienceWorld,
respectively, while Web denotes the evaluation split of WebShop.
All results are averaged over three evaluation runs.}
\label{tab:annealing-1p5b-7b}
\small
\renewcommand{\arraystretch}{1.05}
\setlength{\tabcolsep}{6pt}

\begin{tabular*}{\linewidth}{@{\extracolsep{\fill}}lrrrr@{}}
\toprule
Schedule & ALF & Web & Sci. & Avg. \\
\midrule
\multicolumn{5}{c}{\textit{Different decay shapes, $N=80$}} \\
\midrule
Linear
& \underline{77.86} & 68.75 & \underline{44.01} & \underline{63.54} \\
Cosine
& 71.35 & \underline{71.09} & \underline{44.01} & 62.15 \\
Step
& 69.79 & 65.10 & 36.72 & 57.20 \\

\midrule
\multicolumn{5}{c}{\textit{Linear decay with varying $N$}} \\
\midrule
Linear, $N=80$
& 77.86 & 68.75 & 44.01 & 63.54 \\
Linear, $N=60$
& 76.30 & 68.49 & 45.05 & 63.28 \\
Linear, $N=40$
& 73.96 & \underline{75.00} & 29.69 & 59.55 \\
Linear, $N=20$
& \underline{83.85} & 73.96 & \underline{47.14} & \underline{68.32} \\
Linear, $N=52/29/29$
& 82.55 & 65.89 & 43.23 & 63.89 \\
\midrule
\textbf{\method{}}
& \textbf{84.38} & \textbf{76.30} & \textbf{48.44} & \textbf{69.70} \\
\bottomrule
\end{tabular*}
\end{minipage}
\end{table}

\paragraph{\method{} is robust to the choice of decay shape.}
To isolate the effect of gap-adaptive weighting, we compare \method{} with
linear, cosine, and step decay schedules. All methods use a
Qwen2.5-1.5B teacher and a Qwen2.5-7B student, with schedule definitions provided in
Appendix~\ref{app:scheduling-ablations}. As shown in
Table~\ref{tab:annealing-1p5b-7b}, \method{} achieves the highest success
rate on all three datasets, improving the average success rate by
6.16\%, 7.55\%, and 12.50\% over linear, cosine, and step decay,
respectively. The consistent gains across different decay shapes suggest
that the improvement does not arise from a particular functional form of
weight decay.

\paragraph{Gap-adaptive scheduling avoids the need for a manually tuned horizon.}
We further sweep the horizon $N$ of the linear schedule to examine whether a
well-tuned fixed schedule can match the adaptive strategy. The results show
substantial sensitivity to the choice of $N$, with the best horizon varying
across datasets. Although $N=20$ achieves an average success rate comparable
to \method{}, this setting is identified only through the sweep and does not
transfer consistently across tasks. Moreover, a linear schedule with $N=52/29/29$, calibrated to match \method{}'s teacher-withdrawal steps across ALFWorld, WebShop, and ScienceWorld, still falls substantially behind \method{}. These results suggest that the advantage of \method{} lies
not simply in choosing when to terminate teacher guidance, but in adapting
the OPD weight to the student's evolving capability.

\subsection{Training Efficiency}

\begin{figure}[t]
    \centering
    \includegraphics[width=\linewidth]{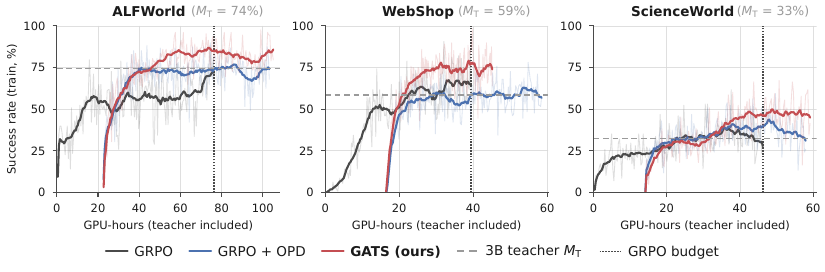}
    \caption{Training success versus GPU-hours for the Qwen2.5-3B$\rightarrow$7B configuration, including teacher costs.
Vertical dotted lines mark the GRPO 150-update budgets; horizontal dashed lines indicate the teacher reference $M_T$.}
    \label{fig:alfworld-gpu-hours}
\end{figure}

\begin{table}[H]
  \centering
  \caption{%
    Smoothed training success rate (\%) at the GRPO 150-update compute budget for each
    environment. $M_T$ is the teacher's training-split success rate used as the
    withdrawal reference.
    \textbf{Bold} indicates the best method in each environment.
  }
  \label{tab:iso-budget}
  \renewcommand{\arraystretch}{0.95}
  \begin{tabular}{lrccccc}
    \toprule
    Environment   & Budget  & GRPO & GRPO+OPD & \method{} & $M_T$
                  & $\Delta$\,(\method{}$-$GRPO) \\
                  & (GPU-h) &      &          &           &
                  & (pp) \\
    \midrule
    ALFWorld      & 76.4 & 71.9 & 73.6          & \textbf{84.9} & 74.4 & $+$13.1 \\
    WebShop       & 39.5 & 64.0 & 58.8          & \textbf{77.1} & 58.6 & $+$13.2 \\
    ScienceWorld  & 46.3 & 27.8 & 39.7          & \textbf{46.0} & 32.5 & $+$18.2 \\
    \bottomrule
  \end{tabular}
\end{table}

\textbf{\method{} achieves higher training success at reference budgets.}
We examine training efficiency in the Qwen2.5-3B$\rightarrow$7B setting using one node with eight H200 GPUs per run.
GPU-hour accounting includes student training and, for teacher-assisted methods, teacher training and online inference.
For each environment, the reference budget is the cost of 150 GRPO updates. Figure~\ref{fig:alfworld-gpu-hours} plots training success rate against cumulative GPU-hours, while Table~\ref{tab:iso-budget} reports the corresponding 15-update centred moving-average values at the reference budgets.
On WebShop and ScienceWorld, \method{} reaches smoothed training success rates of 77.1\% and 46.0\%, respectively, exceeding both GRPO and GRPO+OPD in these comparisons.

%% file: sections/06_conclusion.tex
\section{Conclusion}
\label{sec:conclusion}

We introduced \method{}, a teacher scheduling strategy for sparse-reward
agentic RL. Its central idea is to treat teacher guidance as temporary
assistance: use the measured teacher--student task-performance gap to
adjust the auxiliary OPD weight, then permanently withdraw the teacher
when the smoothed student success rate reaches the teacher reference.
Experiments on three interactive environments show gains over GRPO
and distillation baselines under matched student rollout budgets,
with students exceeding their smaller, task-trained teachers.
Controlled schedule comparisons support adapting guidance to task
progress rather than prescribing its duration in advance.
Future work could explore task-specific scheduling and teacher
reactivation under changing task distributions.

\section*{Reproducibility Statement}
The learning objectives and scheduling rule are specified in
Sections~3 and~4.
Appendix~A reports the evaluation protocol, training configurations,
baseline implementations, and pseudocode.
Appendix~B provides the checkpoint-diagnostic data, and Appendix~C
contains the interaction templates.

\subsection*{AI Use Statement}
Generative AI tools assisted with manuscript revision, narrative and naming
discussions, literature lookup, checks of notation and internal consistency,
and LaTeX and figure preparation. AI-generated suggestions also informed
discussion of methodological caveats and experimental interpretation.
The authors are responsible for the final manuscript and its scientific claims.

%% file: sections/A_implementation_details.tex
\section{Implementation Details}
\label{app:implementation-details}

This appendix describes the benchmark and evaluation protocols,
model and training configurations, and baseline implementations
used in our experiments.

\subsection{Benchmarks and Evaluation}
\label{app:env-protocol}

\paragraph{Benchmarks and splits.}
We evaluate on ALFWorld~\citep{shridhar2020alfworld},
WebShop~\citep{yao2022webshop}, and
ScienceWorld~\citep{wang2022scienceworld}.
For ALFWorld, \texttt{valid\_seen} and \texttt{valid\_unseen}
serve as the in-distribution (ID) and out-of-distribution (OOD)
evaluation splits, respectively.
For ScienceWorld, the training and ID evaluation pools contain
seen tasks, whereas the OOD evaluation pool contains unseen tasks.
WebShop uses a single evaluation split.
Table~\ref{tab:app-task-pools} reports the task-pool sizes.
These sizes are distinct from the number of episodes used in
each evaluation.

\begin{table}[H]
    \centering
    \caption{Task-pool sizes for training and evaluation.
    Dashes indicate splits not used in the reported evaluation.}
    \label{tab:app-task-pools}
    \small
    \setlength{\tabcolsep}{9pt}
    \renewcommand{\arraystretch}{1.0}
    \begin{tabular}{@{}lrrrr@{}}
        \toprule
        Environment & Train & ID & OOD & Eval \\
        \midrule
        ALFWorld     & 3{,}553 & 140 & 134 & -- \\
        WebShop      & 6{,}410 & --  & --  & 500 \\
        ScienceWorld & 3{,}322 & 1{,}661 & 1{,}684 & -- \\
        \bottomrule
    \end{tabular}
\end{table}

\paragraph{Interaction and rewards.}
Episodes are limited to 50 turns in ALFWorld, 15 turns in WebShop,
and 30 turns in ScienceWorld.
ScienceWorld additionally imposes a simulator budget of
100 internal ticks.
An episode is counted as successful only if the environment's
task-completion criterion is satisfied within the applicable
interaction budgets.
Training uses task-completion rewards and, where applicable,
environment-specific invalid-action penalties.
Evaluation success is determined solely by task completion
within these budgets; training-time invalid-action penalties
do not enter the reported success-rate metric.

\paragraph{Evaluation and metrics.}
For each trained policy, we evaluate the final checkpoint three
times, using 128 episodes per evaluation, and report the mean
success rate (SR).
Prompt-only models follow the same evaluation protocol.
Evaluation uses a sampling temperature of $0.4$ and
top-$p=1.0$.
All three repetitions use the same checkpoint.
Avg.~SR is the unweighted mean of five benchmark--split metrics:
ALFWorld ID and OOD, WebShop evaluation, and ScienceWorld ID
and OOD.

\subsection{Models and Training}
\label{app:training-config}

\paragraph{Models and teacher preparation.}
All policies are initialized from the Qwen2.5-Instruct
family~\citep{Yang2024Qwen25TR}.
The teachers are Qwen2.5-1.5B-Instruct and
Qwen2.5-3B-Instruct, and the students are
Qwen2.5-7B-Instruct and Qwen2.5-14B-Instruct.
We evaluate three teacher--student configurations:
$1.5\mathrm{B}\!\rightarrow\!7\mathrm{B}$,
$1.5\mathrm{B}\!\rightarrow\!14\mathrm{B}$, and
$3\mathrm{B}\!\rightarrow\!7\mathrm{B}$.
Each teacher is first trained with GRPO on the corresponding
environment and then frozen throughout student training.

\paragraph{Shared training configuration.}
All trained student methods are run for 150 updates.
Within each environment and student size, we hold the training
data, student rollout budget, and common optimization
hyperparameters fixed across methods.
Table~\ref{tab:app-hyperparams} summarizes the shared training
hyperparameters, while Table~\ref{tab:app-env-hyperparams}
reports environment-specific settings.

\begin{table}[htbp]
\centering
\footnotesize
\caption{Core training hyperparameters shared across student runs.}
\label{tab:app-hyperparams}
\setlength{\tabcolsep}{5pt}
\renewcommand{\arraystretch}{1.05}
\begin{tabular}{@{}ll@{}}
\toprule
Setting & Value \\
\midrule
Training updates & 150 \\
GRPO group size & 8 rollouts per prompt \\
Prompts per batch & 16 \\
Optimizer & AdamW \\
Learning rate & $1\times10^{-6}$ \\
Weight decay & $0.01$ \\
GRPO clip ratio & $0.2$ \\
Training sampling & temperature $1.0$, top-$p=1.0$ \\
\bottomrule
\end{tabular}
\end{table}

\begin{table}[htbp]
\centering
\footnotesize
\caption{Environment-specific training settings.}
\label{tab:app-env-hyperparams}
\setlength{\tabcolsep}{4pt}
\renewcommand{\arraystretch}{1.05}
\begin{tabular}{@{}lccc@{}}
\toprule
Setting & ALFWorld & WebShop & ScienceWorld \\
\midrule
Max prompt length & 2{,}048 & 4{,}096 & 6{,}000 \\
Max response length & 512 & 1{,}024 & 1{,}024 \\
PPO mini-batch size (configuration) & 256 & 64 & 256 \\
Discount $\gamma$ & 0.95 & 1.0 & 1.0 \\
Max turns & 50 & 15 & 30 \\
\bottomrule
\end{tabular}
\end{table}

\subsection{Baseline Implementations}
\label{app:baselines}

We compare \method{} with five baselines.
All trained baselines follow the shared student-training protocol
described in Appendix~\ref{app:training-config}.
The three distillation-based baselines combine GRPO with OPD.

\paragraph{Prompt-only.}
The instruction-tuned student is evaluated without additional
training, using the evaluation protocol described in
Appendix~\ref{app:env-protocol}.

\paragraph{GRPO.}
The student is trained with GRPO without teacher
supervision~\citep{shao2024deepseekmath}.

\paragraph{GRPO $+$ OPD.}
The student is trained with the GRPO objective and the OPD
objective in \eqref{eq:opd-loss}, using a fixed OPD weight
$\lambda_t=1$ throughout training.

\paragraph{ATOD.}
We use the full ATOD method~\citep{tan2026atod}, including its
annealed OPD--RL schedule and Turn-level
Disagreement--Uncertainty Reweighting (T-DUR).

\paragraph{SOD.}
SOD~\citep{zhong2026sod} applies step-level divergence-based
reweighting to OPD without permanently withdrawing teacher
supervision.

\subsection{Scheduling Ablations}
\label{app:scheduling-ablations}

\paragraph{Fixed-time schedules.}
To isolate the effect of the scalar OPD schedule, we hold the teacher,
student, OPD objective, student rollout budget, and all other training
settings fixed, and vary only the scalar OPD weight.
For a horizon $N$ and update indices $t=1,\ldots,150$, we consider
\begin{subequations}
\label{eq:fixed-schedules}
\begin{align}
\lambda_t^{\mathrm{linear}}
&= \max\left\{1-\frac{t}{N},\,0\right\},
\label{eq:linear-schedule}
\\
\lambda_t^{\mathrm{cosine}}
&= \frac{1}{2}
\left[1+\cos\left(\pi\frac{\min\{t,N\}}{N}\right)\right],
\label{eq:cosine-schedule}
\\
\lambda_t^{\mathrm{step}}
&= \mathbf{1}\{t\leq N\}.
\label{eq:step-schedule}
\end{align}
\end{subequations}
The linear and cosine schedules first reach zero at update $N$.
The step schedule retains unit weight through update $N$ and switches
to GRPO alone at update $N+1$.
All three schedules use GRPO alone whenever their OPD weight is zero.

In the decay-shape comparison, we use $N=80$.
For linear decay, we additionally evaluate $N\in\{20,40,60,80\}$.
We also report an environment-specific matched setting with
$N=52,29,29$ for ALFWorld, WebShop, and ScienceWorld, respectively,
to match the observed \method{} withdrawal steps in this configuration.

These schedules depend only on elapsed training updates, whereas
\method{} uses teacher--student success-rate feedback.

\subsection{Gap-Adaptive Teacher Scheduling}
\label{app:method-impl}

\paragraph{Reference and lagged feedback.}
We set $K=5$ for both the teacher reference and the student success-rate
estimate. The teacher reference $M_{\mathrm{T}}$ is computed from the final
five training-time success-rate measurements according to \eqref{eq:teacher-reference}; it is not the held-out teacher score in
Table~\ref{tab:main-results}.
Before student update $t$, $M_{\mathrm{S},t}$ averages the available
success-rate measurements from at most the preceding $K$ student updates.
An empty window gives $M_{\mathrm{S},1}=0$ and hence an initial OPD weight
of one.

\paragraph{Weighting and permanent withdrawal.}
The OPD weight uses only measurements from preceding updates.
Permanent withdrawal is triggered when $M_{\mathrm{S},t}\geq M_{\mathrm{T}}$ before update $t$, after which teacher guidance is permanently disabled and training proceeds with GRPO alone, as specified in Algorithm~\ref{alg:phase-out}. The OPD advantage and loss follow Eqs.~(\ref{eq:opd-advantage}) and~(\ref{eq:opd-loss}), respectively.

\begin{algorithm}[t]
\caption{\method{}: Gap-Adaptive Teacher Scheduling}
\label{alg:phase-out}
\small
\begin{algorithmic}[1]
\Require Frozen teacher $\pi_{\mathrm{T}}$, student $\pi_\theta$,
reference $M_{\mathrm{T}}>0$ from \eqref{eq:teacher-reference}
\Require Task distribution $\mathcal{D}$, group size $G$,
window size $K$, student updates $T$
\Ensure Trained student policy $\pi_\theta$
\State Initialize success-rate queue $Q\gets[\,]$ and $d_0\gets0$
\For{$t=1,\ldots,T$}
    \State $d_t\gets d_{t-1}$
    \If{$Q$ is empty}
        \State $M_{\mathrm{S},t}\gets0$
        \Comment{Initial weight: $\lambda_1=1$}
    \Else
        \State $M_{\mathrm{S},t}\gets\operatorname{mean}(Q)$
    \EndIf
    \If{$M_{\mathrm{S},t}\geq M_{\mathrm{T}}$}
        \State $d_t\gets1$ \Comment{Permanent withdrawal}
    \EndIf
    \State $\lambda_t\gets\max\{1-M_{\mathrm{S},t}/M_{\mathrm{T}},\,0\}$
    \State Sample tasks from $\mathcal{D}$ and $G$ student trajectories per task
    \State Compute returns, rollout success rate $m_{\mathrm{S},t}$,
    and $\mathcal{L}_{\mathrm{GRPO}}$
    \If{$d_t=0$}
        \State Query the teacher on student-generated contexts
        \State Compute $\mathcal{L}_{\mathrm{OPD}}$ using \eqref{eq:opd-loss}
        \State Update the student using \eqref{eq:full-loss}
    \Else
        \State Skip teacher inference and update the student with GRPO only
    \EndIf
    \State Append $m_{\mathrm{S},t}$ to $Q$; retain at most the last $K$ entries
\EndFor
\end{algorithmic}
\end{algorithm}

%% file: sections/B_additional_experiments.tex
\section{Additional Experimental Analysis}
\label{app:additional-analysis}

\subsection{Task-Performance-Gap Diagnostic}
\label{app:gap-diagnostic}

We fix a GRPO-trained Qwen2.5-1.5B teacher and select eight
increasingly capable Qwen2.5-3B-Instruct student checkpoints on
ALFWorld, using a separate 1.5B$\rightarrow$3B configuration from
the main comparison.
Starting from each checkpoint, we run matched 15-update continuations
with GRPO alone and GRPO+OPD with a fixed OPD weight, under the same
student rollout budget.
Table~\ref{tab:gap-diagnostic} provides the data for the gap--utility
diagnostic in the main text.

\begin{table}[htbp]
    \centering
    \small
    \caption{Task-performance-gap diagnostic on ALFWorld.
    The gap is teacher SR minus student SR before the continuation.
    OPD gain is the final SR difference between the matched
    GRPO+OPD and GRPO continuations.
    Both quantities are reported in percentage points (pp).}
    \label{tab:gap-diagnostic}
    \begin{tabular}{@{}cc@{}}
        \toprule
        Teacher--student gap (pp) & OPD gain over GRPO (pp) \\
        \midrule
        $+41.3$ & $+14.1$ \\
        $+31.7$ & $+6.0$ \\
        $+19.0$ & $+5.2$ \\
        $+14.3$ & $+3.4$ \\
        $+9.6$  & $+2.2$ \\
        $-0.8$  & $-3.0$ \\
        $-4.0$  & $-11.7$ \\
        $-9.9$  & $-20.8$ \\
        \bottomrule
    \end{tabular}
\end{table}

Across these continuations, the gain from OPD decreases as the gap
narrows and becomes negative after the student overtakes the teacher.
This diagnostic motivates gap-adaptive supervision but does not
establish a universally optimal withdrawal threshold.

%% file: sections/C_prompts.tex
\section{Prompts and Interaction Templates}
\label{app:prompts}

Each environment uses a history-free template on the first step of an episode
and a history-augmented template on every subsequent step. Both templates
require the model response to contain a reasoning segment between
\lstinline!<think>! and \lstinline!</think>!, followed by an action between
\lstinline!<action>! and \lstinline!</action>!. The history-augmented template
additionally includes the number of previously completed steps and the most
recent \lstinline!history_length! observation--action pairs.

We reproduce the history-augmented template for each environment verbatim
below. The first-step template is identical except that it omits the sentence
reporting the prior step count and recent interaction history.

\subsection{ALFWorld}

\begin{promptbox}{History-Augmented Prompt Template}
You are an expert agent operating in the ALFRED Embodied Environment. Your task is to: {task_description}
Prior to this step, you have already taken {step_count} step(s). Below are the most recent {history_length} observations and the corresponding actions you took: {action_history}
You are now at step {current_step} and your current observation is: {current_observation}
Your admissible actions of the current situation are: [{admissible_actions}].

Now it's your turn to take an action.
You should first reason step-by-step about the current situation. This reasoning process MUST be enclosed within <think> </think> tags.
Once you've finished your reasoning, you should choose an admissible action for current step and present it within <action> </action> tags.
\end{promptbox}

\subsection{WebShop}

\begin{promptbox}{History-Augmented Prompt Template}
You are an expert autonomous agent operating in the WebShop e-commerce environment.
Your task is to: {task_description}.
Prior to this step, you have already taken {step_count} step(s). Below are the most recent {history_length} observations and the corresponding actions you took: {action_history}
You are now at step {current_step} and your current observation is: {current_observation}.
Your admissible actions of the current situation are:
[
{available_actions}
].

Now it's your turn to take one action for the current step.
You should first reason step-by-step about the current situation, then think carefully which admissible action best advances the shopping goal. This reasoning process MUST be enclosed within <think> </think> tags.
Once you've finished your reasoning, you should choose an admissible action for current step and present it within <action> </action> tags.
\end{promptbox}

\subsection{ScienceWorld}

\begin{promptbox}{History-Augmented Prompt Template}
You are an expert agent operating in the ScienceWorld environment, which is a text-based virtual environment centered around accomplishing tasks from the elementary science curriculum.
Your current task is: {task_description}
Prior to this step, you have already taken {step_count} step(s). Below are the most recent {history_length} observations and the corresponding actions you took: {action_history}
You are now at step {current_step} and your current observation is: {current_observation}
Your admissible actions of the current situation are:
{admissible_actions}

Now it's your turn to take an action.
You should first reason step-by-step about the current situation. This reasoning process MUST be enclosed within <think> </think> tags.
Once you've finished your reasoning, you should choose an appropriate action for the current step and present it within <action> </action> tags.
\end{promptbox}

%% file: references.bib
@article{zhou2024archer,
  title={Archer: Training language model agents via hierarchical multi-turn rl},
  author={Zhou, Yifei and Zanette, Andrea and Pan, Jiayi and Levine, Sergey and Kumar, Aviral},
  journal={arXiv preprint arXiv:2402.19446},
  year={2024}
}

@article{bai2024digirl,
  title={Digirl: Training in-the-wild device-control agents with autonomous reinforcement learning},
  author={Bai, Hao and Zhou, Yifei and Cemri, Mert and Pan, Jiayi and Suhr, Alane and Levine, Sergey and Kumar, Aviral},
  journal={Advances in Neural Information Processing Systems},
  volume={37},
  pages={12461--12495},
  year={2024}
}

@article{xi2025agentgym,
  title={Agentgym-rl: Training llm agents for long-horizon decision making through multi-turn reinforcement learning},
  author={Xi, Zhiheng and Huang, Jixuan and Liao, Chenyang and Huang, Baodai and Guo, Honglin and Liu, Jiaqi and Zheng, Rui and Ye, Junjie and Zhang, Jiazheng and Chen, Wenxiang and others},
  journal={arXiv preprint arXiv:2509.08755},
  year={2025}
}

@article{shao2024deepseekmath,
  title={Deepseekmath: Pushing the limits of mathematical reasoning in open language models},
  author={Shao, Zhihong and Wang, Peiyi and Zhu, Qihao and Xu, Runxin and Song, Junxiao and Bi, Xiao and Zhang, Haowei and Zhang, Mingchuan and Li, YK and Wu, Yang and others},
  journal={arXiv preprint arXiv:2402.03300},
  year={2024}
}

@article{guo2025deepseekr1,
  title={DeepSeek-R1 incentivizes reasoning in LLMs through reinforcement learning},
  author={Guo, Daya and Yang, Dejian and Zhang, Haowei and Song, Junxiao and Wang, Peiyi and Zhu, Qihao and Xu, Runxin and Zhang, Ruoyu and Ma, Shirong and Bi, Xiao and others},
  journal={Nature},
  volume={645},
  number={8081},
  pages={633--638},
  year={2025},
  publisher={Nature Publishing Group UK London}
}

@article{yan2025luffy,
  title={Learning to reason under off-policy guidance},
  author={Yan, Jianhao and Li, Yafu and Hu, Zican and Wang, Zhi and Cui, Ganqu and Qu, Xiaoye and Cheng, Yu and Zhang, Yue},
  journal={Advances in Neural Information Processing Systems},
  volume={38},
  pages={117157--117186},
  year={2026}
}

@article{zheng2025selective,
  title={Act only when it pays: Efficient reinforcement learning for llm reasoning via selective rollouts},
  author={Zheng, Haizhong and Zhou, Yang and Bartoldson, Brian and Kailkhura, Bhavya and Lai, Fan and Zhao, Jiawei and Chen, Beidi},
  journal={Advances in Neural Information Processing Systems},
  volume={38},
  pages={124321--124346},
  year={2026}
}

@article{zhang2025bread,
  title={Bread: Branched rollouts from expert anchors bridge sft \& rl for reasoning},
  author={Zhang, Xuechen and Huang, Zijian and Li, Yingcong and Ni, Chenshun and Chen, Jiasi and Oymak, Samet},
  journal={Advances in Neural Information Processing Systems},
  volume={38},
  pages={96726--96752},
  year={2026}
}

@article{huang2026prefixrft,
  title={Blending supervised and reinforcement fine-tuning with prefix sampling, 2025},
  author={Huang, Zeyu and Cheng, Tianhao and Qiu, Zihan and Wang, Zili and Xu, Yinghui and Ponti, Edoardo M and Titov, Ivan},
  journal={URL https://arxiv. org/abs/2507.01679},
  year={2025}
}

@inproceedings{zhang2026chord,
  title={On-policy rl meets off-policy experts: Harmonizing supervised fine-tuning and reinforcement learning via dynamic weighting},
  author={Zhang, Wenhao and Xie, Yuexiang and Sun, Yuchang and Chen, Yanxi and Wang, Guoyin and Li, Yaliang and Ding, Bolin and Zhou, Jingren},
  booktitle={International Conference on Learning Representations},
  volume={2026},
  pages={120693--120726},
  year={2026}
}

@article{chu2025sftmemorizes,
  title={Sft memorizes, rl generalizes: A comparative study of foundation model post-training},
  author={Chu, Tianzhe and Zhai, Yuexiang and Yang, Jihan and Tong, Shengbang and Xie, Saining and Schuurmans, Dale and Le, Quoc V and Levine, Sergey and Ma, Yi},
  journal={arXiv preprint arXiv:2501.17161},
  year={2025}
}

@inproceedings{jiang2026mentor,
  title={Selective expert guidance for effective and diverse exploration in reinforcement learning of llms},
  author={Jiang, Zishang and Han, Jinyi and Wang, Xinyi and Jiang, Sihang and Dai, Zhaoqian and Shuguang, Ma and Yu, Fei and Liang, Jiaqing and Xiao, Yanghua and others},
  booktitle={International Conference on Learning Representations},
  volume={2026},
  pages={62980--63006},
  year={2026}
}

@article{li2026sequential,
  title={Sequential Beats Joint: On the Interplay between On-Policy Distillation and RLVR},
  author={Li, Boyan and Chen, Bingsen and Yang, Chenghao and Nie, Ping and Zhao, Chen and Ye, Xi},
  journal={arXiv preprint arXiv:2609.04108},
  year={2026}
}

@article{shridhar2020alfworld,
  title={Alfworld: Aligning text and embodied environments for interactive learning},
  author={Shridhar, Mohit and Yuan, Xingdi and C{\^o}t{\'e}, Marc-Alexandre and Bisk, Yonatan and Trischler, Adam and Hausknecht, Matthew},
  journal={arXiv preprint arXiv:2010.03768},
  year={2020}
}

@article{yao2022webshop,
  title={Webshop: Towards scalable real-world web interaction with grounded language agents},
  author={Yao, Shunyu and Chen, Howard and Yang, John and Narasimhan, Karthik},
  journal={Advances in Neural Information Processing Systems},
  volume={35},
  pages={20744--20757},
  year={2022}
}

@inproceedings{wang2022scienceworld,
  title={Scienceworld: Is your agent smarter than a 5th grader?},
  author={Wang, Ruoyao and Jansen, Peter and C{\^o}t{\'e}, Marc-Alexandre and Ammanabrolu, Prithviraj},
  booktitle={Proceedings of the 2022 Conference on Empirical Methods in Natural Language Processing},
  pages={11279--11298},
  year={2022}
}

@misc{Yang2024Qwen25TR,
      title={Qwen2.5 Technical Report}, 
      author={Qwen and An Yang and Baosong Yang and Beichen Zhang and Binyuan Hui and Bo Zheng and Bowen Yu and Chengyuan Li and Dayiheng Liu and Fei Huang and Haoran Wei and Huan Lin and Jian Yang and Jianhong Tu and Jianwei Zhang and Jianxin Yang and Jiaxi Yang and Jingren Zhou and Junyang Lin and Kai Dang and Keming Lu and Keqin Bao and Kexin Yang and Le Yu and Mei Li and Mingfeng Xue and Pei Zhang and Qin Zhu and Rui Men and Runji Lin and Tianhao Li and Tianyi Tang and Tingyu Xia and Xingzhang Ren and Xuancheng Ren and Yang Fan and Yang Su and Yichang Zhang and Yu Wan and Yuqiong Liu and Zeyu Cui and Zhenru Zhang and Zihan Qiu},
      year={2025},
      eprint={2412.15115},
      archivePrefix={arXiv},
      primaryClass={cs.CL},
      url={https://arxiv.org/abs/2412.15115}, 
}

@inproceedings{ji2026tree,
  title={Tree search for llm agent reinforcement learning},
  author={Ji, Yuxiang and Ma, Ziyu and Wang, Yong and Chen, Guanhua and Chu, Xiangxiang and Wu, Liaoni},
  booktitle={International Conference on Learning Representations},
  volume={2026},
  pages={87362--87388},
  year={2026}
}

@article{liu2026uft,
  title={Uft: Unifying supervised and reinforcement fine-tuning},
  author={Liu, Mingyang and Farina, Gabriele and Ozdaglar, Asuman},
  journal={Advances in Neural Information Processing Systems},
  volume={38},
  pages={101347--101383},
  year={2026}
}

@inproceedings{feng2026retool,
  title={Retool: Reinforcement learning for strategic tool use in llms},
  author={Feng, Jiazhan and Huang, Shijue and Qu, Xingwei and Zhang, Ge and Qin, Yujia and Zhong, Baoquan and Jiang, Chengquan and Chi, Jinxin and Zhong, Wanjun},
  booktitle={International Conference on Learning Representations},
  volume={2026},
  pages={37909--37926},
  year={2026}
}

@article{jin2025search,
  title={Search-r1: Training llms to reason and leverage search engines with reinforcement learning},
  author={Jin, Bowen and Zeng, Hansi and Yue, Zhenrui and Yoon, Jinsung and Arik, Sercan and Wang, Dong and Zamani, Hamed and Han, Jiawei},
  journal={arXiv preprint arXiv:2503.09516},
  year={2025}
}

@article{wang2025ragen,
  title={Ragen: Understanding self-evolution in llm agents via multi-turn reinforcement learning},
  author={Wang, Zihan and Wang, Kangrui and Wang, Qineng and Zhang, Pingyue and Li, Linjie and Yang, Zhengyuan and Jin, Xing and Yu, Kefan and Nguyen, Minh Nhat and Liu, Licheng and others},
  journal={arXiv preprint arXiv:2504.20073},
  year={2025}
}

@article{yu2026dapo,
  title={Dapo: An open-source llm reinforcement learning system at scale},
  author={Yu, Qiying and Zhang, Zheng and Zhu, Ruofei and Yuan, Yufeng and Zuo, Xiaochen and Yue, Yu and Dai, Weinan and Fan, Tiantian and Liu, Gaohong and Liu, Lingjun and others},
  journal={Advances in Neural Information Processing Systems},
  volume={38},
  pages={113222--113244},
  year={2026}
}

@inproceedings{agarwal2024policy,
  title={On-policy distillation of language models: Learning from self-generated mistakes},
  author={Agarwal, Rishabh and Vieillard, Nino and Zhou, Yongchao and Stanczyk, Piotr and Ramos Garea, Sabela and Geist, Matthieu and Bachem, Olivier},
  booktitle={International Conference on Learning Representations},
  volume={2024},
  pages={21246--21263},
  year={2024}
}

@article{tan2026atod,
  title={ATOD: Annealed Turn-aware On-policy Distillation for Multi-turn Autonomous Agents},
  author={Tan, Qitai and Zong, Zefang and Li, Yang and Chen, Peng},
  journal={arXiv preprint arXiv:2606.27814},
  year={2026}
}

@article{zhong2026sod,
  title={Sod: Step-wise on-policy distillation for small language model agents},
  author={Zhong, Qiyong and Zheng, Mao and Song, Mingyang and Lin, Xin and Sun, Jie and Jiang, Houcheng and Wang, Xiang and Fang, Junfeng},
  journal={arXiv preprint arXiv:2605.07725},
  year={2026}
}

@article{li2026unifying,
  title={Unifying group-relative and self-distillation policy optimization via sample routing},
  author={Li, Gengsheng and Yang, Tianyu and Fang, Junfeng and Song, Mingyang and Zheng, Mao and Guo, Haiyun and Zhang, Dan and Wang, Jinqiao and Chua, Tat-Seng},
  journal={arXiv preprint arXiv:2604.02288},
  year={2026}
}

@article{ding2026hdpo,
  title={Hdpo: Hybrid distillation policy optimization via privileged self-distillation},
  author={Ding, Ken},
  journal={arXiv preprint arXiv:2603.23871},
  year={2026}
}

@article{thinkingmachinlab-opd,
  author = {Kevin Lu and Thinking Machines Lab},
  title = {On-Policy Distillation},
  journal = {Thinking Machines Lab: Connectionism},
  year = {2025},
  note = {https://thinkingmachines.ai/blog/on-policy-distillation},
  doi = {10.64434/tml.20251026},
}
